\documentclass[letterpaper, 10 pt, conference]{ieeeconf}  

\IEEEoverridecommandlockouts                              

\usepackage{graphics} 
\usepackage{epsfig} 
\usepackage{mathptmx} 
\usepackage{times} 

\usepackage{amsmath,amssymb,amsfonts}
\usepackage{algorithmic}
\usepackage{graphicx}
\usepackage{algorithm,algorithmic}
\usepackage{hyperref}
\hypersetup{hidelinks=true}
\usepackage{textcomp}
\usepackage{amsmath}
\usepackage{flushend}
\usepackage{booktabs}
\usepackage{makecell}
\usepackage{threeparttable}
\usepackage{multirow}
\usepackage{gensymb} 
\usepackage{color} 
\usepackage[sorting=none, style=ieee]{biblatex}
\usepackage[T1]{fontenc}

\title{\LARGE \bf
DD-VNB: A Depth-based Dual-Loop Framework for Real-time Visually Navigated Bronchoscopy
}

\author{Qingyao Tian, Huai Liao, Xinyan Huang, Jian Chen, Zihui Zhang, Bingyu Yang, \\ Sebastien Ourselin and Hongbin Liu
\thanks{
Qingyao Tian, Jian Chen and Bingyu Yang are with Institute of Automation, Chinese Academy of Sciences, Beijing 100190, China, and also with the School of Artificial Intelligence, University of Chinese Academy of Sciences, Beijing 100049, China.
}
\thanks{
Huai Liao, M.D. and Xin-yan Huang, M.D. are with Department of Pulmonary and Critical Care Medicine, The First Affiliated Hospital of Sun Yat-sen University, Guangzhou, Guangdong Province, P.R. China.
}
\thanks{
Zihui Zhang is with Institute of Automation, Chinese Academy of Sciences, Beijing, 100190, China.
}
\thanks{Sebastien Ourselin is with School of Biomedical Engineering and Imaging Sciences, King’s College London, London, UK.
}
\thanks{
Corresponding author: Hongbin Liu is with Institute of Automation, Chinese Academy of Sciences, and with Centre of AI and Robotics, Hong Kong Institute of Science\& Innovation, Chinese Academy of Sciences. He is also affiliated with the School of Biomedical Engineering and Imaging Sciences, King’s College London, UK. (e-mail: liuhongbin@ia.ac.cn).
}}

\begin{document}

\maketitle

\pubid{\begin{minipage}{\textwidth}\ \\[30pt] \centering
	This work has been submitted to the IEEE for possible publication. Copyright may be transferred without notice, after which this version may no longer be accessible.
\end{minipage}}

\begin{abstract}

Real-time 6 DOF localization of bronchoscopes is crucial for enhancing intervention quality. However, current vision-based technologies struggle to balance between generalization to unseen data and computational speed. In this study, we propose a Depth-based Dual-Loop framework for real-time Visually Navigated Bronchoscopy (DD-VNB) that can generalize across patient cases without the need of re-training. The DD-VNB framework integrates two key modules: depth estimation and dual-loop localization. To address the domain gap among patients, we propose a knowledge-embedded depth estimation network that maps endoscope frames to depth, ensuring generalization by eliminating patient-specific textures. The network embeds view synthesis knowledge into a cycle adversarial architecture for scale-constrained monocular depth estimation. For real-time performance, our localization module embeds a fast ego-motion estimation network into the loop of depth registration. The ego-motion inference network estimates the pose change of the bronchoscope in high frequency while depth registration against the pre-operative 3D model provides absolute pose periodically. Specifically, the relative pose changes are fed into the registration process as the initial guess to boost its accuracy and speed. Experiments on phantom and in-vivo data from patients demonstrate the effectiveness of our framework: 1) monocular depth estimation outperforms SOTA, 2) localization achieves an accuracy of Absolute Tracking Error (ATE) of 4.7 $\pm$ 3.17 mm in phantom and 6.49 $\pm$ 3.88 mm in patient data, 3) with a frame-rate approaching video capture speed, 4) without the necessity of case-wise network retraining. The framework's superior speed and accuracy demonstrate its promising clinical potential for real-time bronchoscopic navigation.

\end{abstract}

\begin{figure}[tbp]
\centerline{\includegraphics[width=\columnwidth]{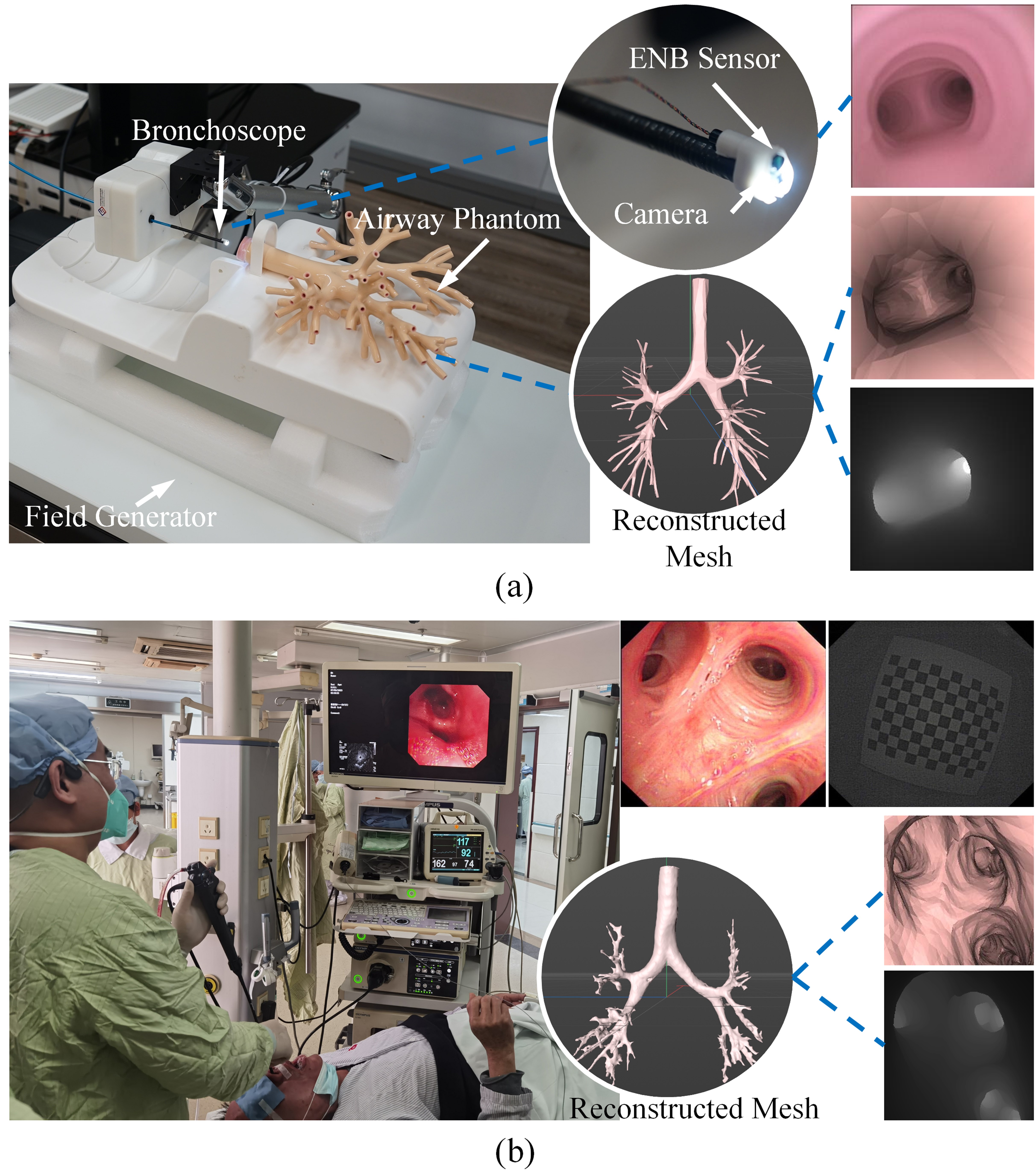}}
\caption{Experimental set up (a) using phantom, (b) by acquiring patient data.}
\label{fig3}
\end{figure}

\section{Introduction}
\label{sec:introduction}
Lung cancer stands as a major health concern on a global scale \cite{siegel2023cancer}. Nevertheless, it is a largely preventable disease and early diagnosis significantly improves patient outcomes \cite{metintacs2023epidemiology}. Bronchoscopy has played a pivotal role in the examination and diagnosis of airway lesions \cite{criner2020interventional}. During bronchoscopy, surgeons navigate a flexible bronchoscope with a distal camera towards nodules identified in pre-operative CT scans based on endoscopic images. The limited field of view of the bronchoscope necessitates extensive clinical experience to localize its position within the airway. Therefore, there is an urgent demand for a bronchoscopic localization method to aid in interventions.

Recent advancements in robotic bronchoscopy \cite{reisenauer2022ion,ho2022feasibility} have highlighted technologies like electromagnetic navigation \cite{folch2020sensitivity} and 3-D shape sensing \cite{shi2016shape} as significant aids. Visually Navigated Bronchoscopy (VNB) emerges as a promising area of study due to its potential for high accuracy and simple setup. However, the limitations of monocular bronchoscope settings and the textureless airway surface hinder the application of traditional SLAM methods \cite{visentini2017deep}. Furthermore, variations in patient airways pose a challenge in developing a streamlined method for generalization \cite{zhao2019generative,sganga2019autonomous}. Consequently, current VNB methods struggle to meet practical application demands. 

The primary limitation of existing VNB methods lies in balancing computational speed and creating a streamlined pipeline to generalize across patients. Existing VNB methods mainly belong to two categories: registration-based \cite{mori2002tracking,deguchi2009selective,shen2019context,banach2021visually} and retrieval-based \cite{zhao2019generative,sganga2019autonomous} localization. registration-based approaches rely on iterative optimization, resulting in limited computational speed, while retrieval-based methods adapt to different patients by case-wise re-training.

In order to address the challenge of balancing speed and generalization, this paper proposes a Depth-based Dual-Loop Framework for Visually Navigated Bronchoscopy (DD-VNB). The methodology involves depth estimation and a dual-loop localization module. The depth estimation of endoscope frames serves as input for the localization module.

To achieve generalization across cases, we propose a knowledge-embeded depth estimation network. Initially, the depth estimation maps endoscope frames to the depth space, serving as input for the pose inference module. This process eliminates patient-specific textures, ensuring the generalization ability of the localization method. 
Moreover, we innovatively incorporates view synthesis into an unpaired image translation framework, leveraging the geometric principles of view synthesis to inform and constrain the learning process. This approach is motivated by earlier research that utilized view synthesis for joint training of unsupervised depth and motion estimation \cite{zhou2017unsupervised}, as well as recent efforts to apply geometry view synthesis in creating realistic surgical scenes \cite{rivoir2021long}.

To achieve real-time speed, during the localization stage, DD-VNB embeds a fast ego-motion estimation network within the loop of depth map registration for real-time performance. The ego-motion inference network estimates the bronchoscope's pose change at high frequency, while depth registration against the pre-operative 3D model provides absolute pose values periodically. Specifically, the relative pose changes are fed into the registration process as the initial guess to boost accuracy and speed.

In summary, our contributions are as follows:

\begin{enumerate}
    \item A generalizable VNB framework for real-time application is presented, which can generalize across different patient cases, eliminating the necessity for re-training.
    \item  We propose a knowledge-embeded depth estimation network that leverages geometry view synthesis for accurate depth estimation from monocular endoscopic images at a specific scale.
    \item A fast localization module is proposed that embeds an efficient ego-motion estimation network within the loop of single-view depth map registration, enabling fast and accurate pose inference.
    \item The proposed framework outperforms SOTA in localization accuracy, as demonstrated by extensive experiments on both phantom and patient data, showcasing its practical effectiveness (Fig. \ref{fig3}).
\end{enumerate}

\section{Related Works}
Recent advances in deep learning (DL) prompts exploration into learning-based VNB \cite{ozyoruk2021endoslam, fried2023landmark,deng2023feature,sganga2019offsetnet,sganga2019autonomous,zhao2019generative,karaoglu2021adversarial,shen2019context,banach2021visually,borrego2023bronchopose}. Ozyoruk et al. \cite{ozyoruk2021endoslam} and Deng et al. \cite{deng2023feature} introduce visual odometry methods to endoscopic videos. Nonetheless, existing deep monocular visual odometry algorithms face challenges such as scale ambiguity and drift, making them less suitable for bronchoscopic applications. OffsetNet \cite{sganga2019offsetnet} employs DL to register between real and rendered images but suffers from low accuracy in unseen areas during training. Their later work, AirwayNet \cite{sganga2019autonomous}, localizes the camera by estimating visible airways and their relative poses, with successful navigation critically dependent on identifying airways. Zhao et al. \cite{zhao2019generative} adopt auxiliary learning to train a global pose regression network \cite{valada2018deep}. However, global pose learning is essentially image retrieval and cannot generalize beyond their training data \cite{sattler2019understanding}.

Several studies \cite{karaoglu2021adversarial,shen2019context,banach2021visually,mathew2020augmenting} explore depth estimation of bronchoscopic images by cycle adversarial networks \cite{zhu2017unpaired} and perform registration \cite{shen2019context,banach2021visually}. Theoretically, with adequate training, by utilizing real bronchoscopic images and unpaired virtual depth maps as inputs, cycle adversarial networks are capable of learning the mapping from the distribution of bronchoscopic images to the distribution of their corresponding depth maps \cite{zhu2017unpaired}. By training an unsupervised conditional cycle adversarial network, Shen et al. \cite{shen2019context} obtain a mapping network from bronchoscopic frames to corresponding depth maps, addressing the problem of visual artifacts. Localization is accomplished by iteratively registering the estimated depth maps to preoperative images. Banach et al. \cite{banach2021visually} use a Three Cycle-Consistent Generative Adversarial Network architecture to estimate image depth and register the generated point cloud to a pre-operative airway model. In these studies, depth-based methods prove robust due to the removal of individual differences in illumination and texture. However, the work mentioned above faces limitations in real-world applications for two primary reasons. Firstly, due to its loose constraint, the cycle adversarial network often achieves biased distribution mapping, resulting in unstable scale between different frames and possible changes in object structure \cite{karaoglu2021adversarial}. Secondly, it concentrates exclusively on depth estimation and relies on traditional algorithms for registering camera poses, which suffer from low update frame rates.

In this paper, we aim to achieve real-time bronchoscope localization with generalization to unseen cases during training. To address the issue of learning scale-informed depth, we introduce view consistency loss and geometry consistency loss, which significantly advances the depth estimation process. Additionally, to ensure real-time localization, we design a fast localization module that incorporates an efficient ego-motion estimation network into the depth registration loop. 

\begin{figure*}[tbp]
\centerline{\includegraphics[width=15cm]{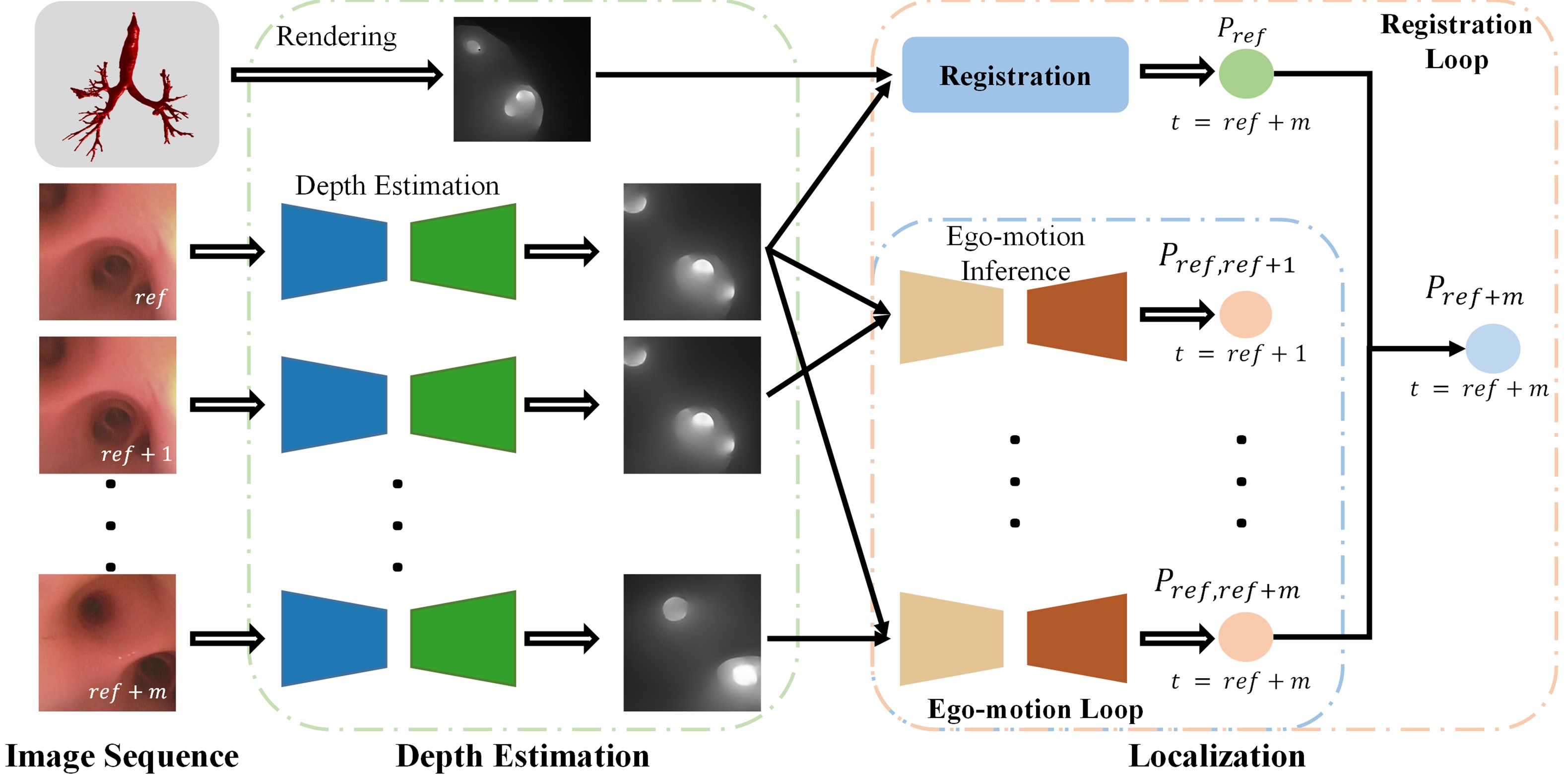}}
\caption{The overview of proposed framework, where $ref$ denotes reference time point. During intervention, we first estimate the incoming bronchoscopic frame’s depth map. Then, a dual-loop diagram is introduced to locate the camera position. The ego-motion loop tracks the camera position by inferring the camera movement between a pair of input depth maps in real time. The registration loop infers global pose by referring to pre-operative airway map and inertially eliminates accumulative error by ego-motion estimation. For the next iteration, $ref+m$ serves as the next reference time point for dual-loop iteration, and $P_{ref+m}$ is considered as the initial value for registration.}
\label{fig1}
\end{figure*}

\section{Method}
Fig. \ref{fig1} shows an overview of the proposed bronchoscopic localization framework. The localization methodology involves depth estimation and localization. Learning-based depth estimation is accomplished based on a adversarial architecture with knowledge embedding as supervision. The localization part consists two algorithms: an ego-motion estimation network that takes two nearby bronchoscopic depth maps as input to infer a 6 DOF relative pose between them; and a depth map registration method mapping a single estimated depth map to the pre-operative airway model, correcting the accumulative error of incremental ego-motion inference.

\subsection{Depth Estimation with View Synthesis as Supervision}
\begin{figure}[tbp]
\centerline{\includegraphics[width=\columnwidth]{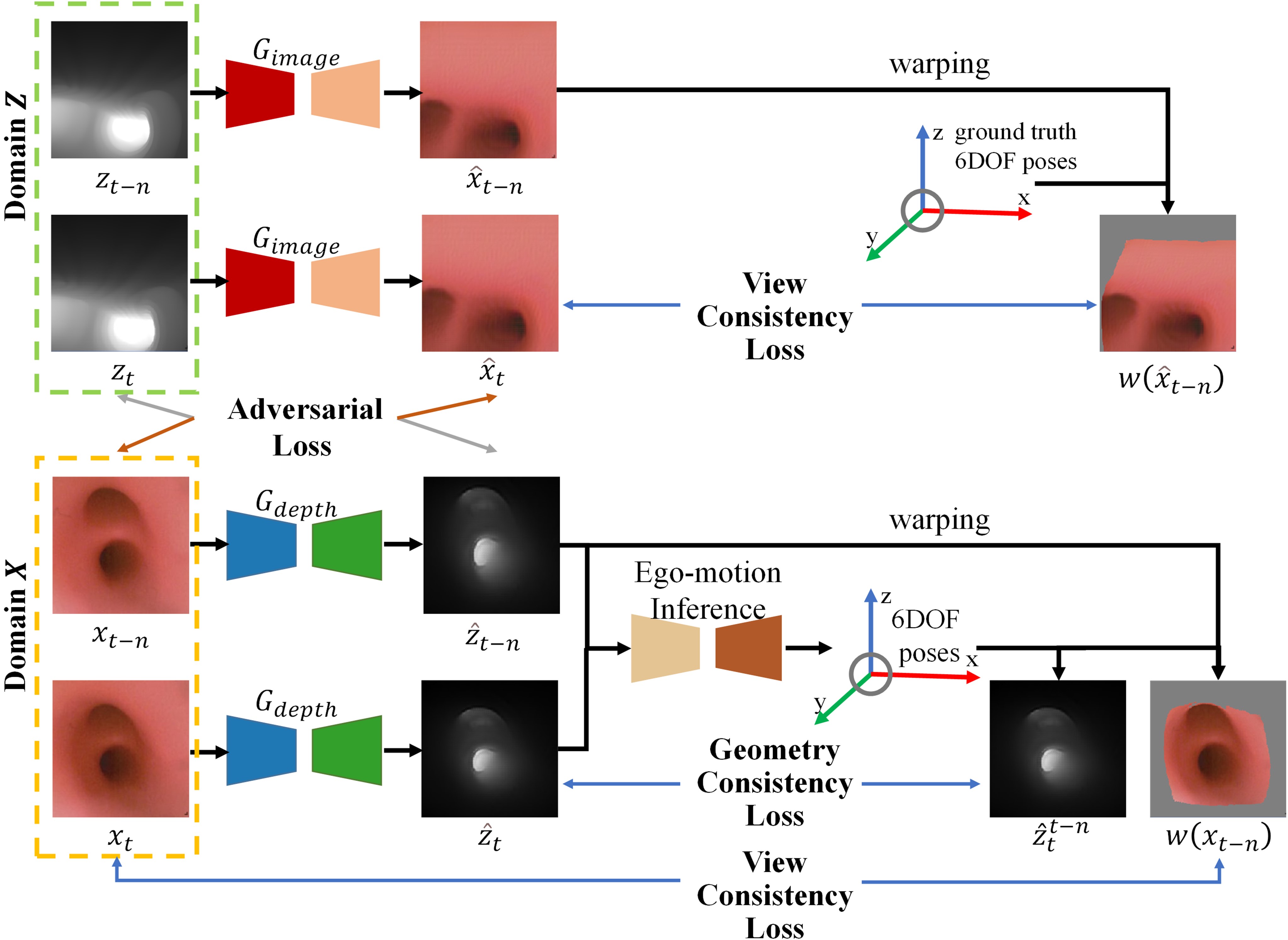}}
\caption{Our depth estimation network training incorporates scale-awareness by combining unpaired image-to-image translation with view synthesis, enforcing view consistency during training. In the \(X \rightarrow Z\) direction (lower half), depth maps \(\hat{z}_t\) and \(\hat{z}_{t-n}\) are generated for frames \(x_t\) and \(x_{t-n}\), and camera motion is inferred by the pretrained ego-motion estimation network. With depth and motion, view-synthesized image \(w\left(x_{t-n}\right)\) and reprojected depth \(\hat{z}_t^{t-n}\) are obtained, enforcing consistency between \(x_t\) and \(w\left(x_{t-n}\right)\), and geometry consistency between \(\hat{z}_t\) and \(\hat{z}_t^{t-n}\). In the \(Z \rightarrow X\) direction (upper half), ground truth pose and depth in virtual bronchoscopy yield \(w\left(\hat{x}_{t-n}\right)\), enforcing view consistency with \(\hat{x}_t\). Adversarial loss in the diagram combines discriminators \(D_{{\text{depth}}}\) and \(D_{{\text{image}}}\).}
\label{fig2}
\end{figure}

The training strategy of the proposed depth estimation module is shown in Fig. \ref{fig2}. Training of our depth estimation network employs an unpaired image-to-image translation framework based on cycle consistency \cite{zhu2017unpaired}. The depth estimation aims to map a bronchoscopic frame $x \in X$ to its depth space by $G_{{depth}}: X \rightarrow Z$, generating its corresponding depth map $\hat{z}_t=G_{{depth}}\left(x_t\right)$. The cycle is completed by reconstructing $\hat{z}_t$ back to the domain $X$ using $G_{{image}}: Z \rightarrow X$. The translation from $Z$ to $X$ is similarly performed. The model enforces cycle consistency for $G_{{depth}}$ and $G_{{image}}$. Finally, discriminators $D_{{depth}}$ and $D_{{image}}$ distinguish between generated/real depth maps and video frames, respectively.

We use the LS-GAN loss \cite{mao2017least} as adversarial loss $L_{adv}$ and L1 losses for $L_{cyc}$ and $L_{iden}$ to guide the networks to learn domain transferring as preserving important structure.

\vspace{0.3cm}
\textbf{View Consistency Loss}: To enforce the network to learn depth estimation with absolute scale, virtual camera poses of input depth maps are also collected from our simulator to impose view consistency between generated bronchoscopic video frames.

Taking depth map $z_{t-n}$ and $z_t$ as input, network $G_{{image}}$ generates bronchoscopic images $\hat{x}_{t-n}$ and $\hat{x}_t$. With ground truth 6 DOF relative camera pose $P_{t-n, t}$ between depth map $z_{t-n}$ and $z_t$, along with camera intrinsic $K$, a homogeneous pixel $p_{t-n} \in \hat{x}_{t-n}$ can be warped to $\hat{p}_t \in \hat{x}_t$ according to view synthesis:

\begin{equation}\label{eq:example}
    \hat{z}_t \hat{p}_t=\boldsymbol{K} \boldsymbol{R}_{t-n, t} \boldsymbol{K}^{-1} z_{t-n} p_{t-n}+\boldsymbol{K} T_{t-n, t},
\end{equation}

\noindent where $T_{t-n, t}$ and $\boldsymbol{R}_{t-n, t}$ are the translation vector and rotation matrix from $t-n$ to $t$.

Differentiable bilinear sampling is applied to $\hat{x}_{t-n}$ according to continuous coordinates $\hat{p}_t$, mapping $\hat{x}_{t-n}$ to warped bronchoscopic image $w\left(\hat{x}_{t-n}\right)$, which is supposed to be in consistent with $\hat{x}_t$. Therefore, we propose to minimize the pixelwise inconsistency between warped frame $w\left(\hat{x}_{t-n}\right)$ and generated frame $\hat{x}_t$. For a pair of generated frames $\hat{x}_{t-n}$ and $\hat{x}_t$, the view consistency loss is defined as

\begin{equation}\label{eq:example}
   L_{r e c_{-} \hat{x}}=\frac{1}{|V|} \sum_{p \in V}\left|w\left(\hat{x}_{t-n}\right)(p)-\hat{x}_t(p)\right|,
\end{equation}

\noindent where $w(\cdot)$ is the warping operator into the pixel space of $\hat{x}_t$ by (1); $V$ stands for valid pixels successfully projected from $\hat{x}_{t-n}$ to the image plane of $\hat{x}_t,|V|$ represents the number of pixels in $V$. By this means, $G_{{image}}$ is enforced to learn the unbiased mapping from depth map to its corresponding bronchoscopic frame. As cycle consistency is enforced, $G_{{depth}}$ would in turn learn the unbiased mapping from input video frames to scale-constrained depth maps.

To further constrain the learning of mapping $G_{{depth}}: X \rightarrow Z$, view synthesis is applied to input bronchoscopic frames $x_{t-n}$ and $x_t$ with their generated depth maps. Although ground truth camera motion between $x_{t-n}$ and $x_t$ is not accessible, as the estimated depth maps $\hat{z}_{t-n}$ and $\hat{z}_t$ are obtained, the inferred relative pose $\hat{P}_{t-n, t}$ could be calculated by ego-motion estimation, which will be discussed in detail in Section III.B. Thus, a satisfying depth map estimation should be informative to recover the camera motion and preserve object structure, which is formulated as a view consistency loss of:

\begin{equation}\label{eq:example}
   L_{r e c_{-} x}=\frac{1}{|V|} \sum_{p \in V}\left|w\left(x_{t-n}\right)(p)-x_t(p)\right|.
\end{equation}

Combining (2) and (3), the complete view consistency loss is defined as:

\begin{equation}\label{eq:example}
   L_{r e c}=\tau_1 L_{r e c_{-} \hat{x}}+\tau_2 L_{r e c_{-} x},
\end{equation}

\noindent where $\tau_1$ and $\tau_2$ are weight terms used to balance losses.

\vspace{0.3cm}
\textbf{Geometry Consistency Loss}: For generated consecutive depth map $\hat{z}_{t-n}$ and $\hat{z}_t$, suppose they conform the same 3D scene structure, the difference of their 3D depth attributes should be minimized. Following \cite{bian2019unsupervised}, we enforce geometry consistency loss on the predicted depth maps so that scale of a depth estimation sequence should all agree with each other, and as a result further constrain the learning of mapping $G_{{depth}}: X \rightarrow Z$.

For generated consecutive depth map $\hat{z}_{t-n}$ and $\hat{z}_t$, the depth inconsistency map $z_{{diff}}$ is defined in \cite{bian2019unsupervised} as:

\begin{equation}\label{eq:example}
   z_{{diff}}=\frac{\left|\hat{z}_t^{t-n}-\hat{z}_t^{\prime}\right|}{\hat{z}_t^{t-n}+\hat{z}_t^{\prime}},
\end{equation}

\noindent where $\hat{z}_t^{t-n}$ is the computed depth map of $\hat{z}_t$ by warping $\hat{z}_{t-n}$ using inferred camera motion $\hat{p}_{t-n, t}$, and $\hat{z}_t^{\prime}$ is the interpolated depth map from generated depth map $\hat{z}_t$. We use $\hat{z}_t^{\prime}$ instead of $\hat{z}_t$ because the warping flow from $\hat{z}_{t-n}$ to $\hat{z}_t^{t-n}$ does not lie on the pixel grid. The depth inconsistency map is normalized by the sum of the two depth maps.
The geometry consistency loss can then be defined as:

\begin{equation}\label{eq:example}
   L_{g c}=\frac{1}{|V|} \sum_{p \in V} z_{{diff}}(p).
\end{equation}

Combining the above loss terms, the overall loss is as follows:

\begin{equation}\label{eq:example}
   \begin{aligned}
L=\beta L_{c y c}+\gamma L_{i d e n} +\delta L_{a d v} + L_{r e c}+\eta L_{g c},
\end{aligned}
\end{equation}

\noindent where $\beta$, $\gamma$, $\delta$ and $\eta$ are weight terms used to balance losses.

\subsection{Dual-loop Localization}
Our localization module integrates an ego-motion estimation network into the loop of depth registration for real-time bronchoscope tracking, enhancing accuracy by using relative pose changes as initial guesses in depth registration against the pre-operative 3D model.

\vspace{0.3cm}
\textbf{Depth Based Ego-motion Inference.}
As depth map and corresponding camera poses are accessible in virtual bronchoscopy, the learning objective is to minimize the difference between predicted transformation $\widehat{P}_{t-n, t}$ and ground truth transformation $P_{t-n, t}$. Therefore, the relative camera pose inference loss is defined as:

\begin{equation}\label{eq:example}
  \begin{aligned}
  L\left(z_{t-n}, z_t\right) & =\left\|T_{t-n, t}-\hat{T}_{t-n, t}\right\|_2\\
& +\omega\left\|r_{t-n, t}-\hat{r}_{t-n, t}\right\|_2,
\end{aligned}
\end{equation}

\noindent where $T_{t-n, t}$ and $\hat{T}_{t-n, t}$ are the ground truth and predicted translation vector respectively. $r_{t-n, t}$ and $\hat{r}_{t-n, t}$ are the ground truth and predicted Euler angles. For data augmentation, depth map $z_{t-n}, z_t, n \in$ $[-5,5]$ is randomly sampled from virtual bronchoscopy sequences for guaranteeing sufficient co-visibility while enhancing motion diversity.

We use FlownetC encoder \cite{ilg2017flownet} to extract image features. Then, five convolutional blocks are utilized to regress a pose vector from extracted features. By using virtual depth maps and poses as training data, our ego-motion network can be deployed directly to estimated depth map of incoming bronchoscopic frames during test time. 

\vspace{0.3cm}
\textbf{Depth Map Registration.}
Relying on ego-motion estimation network alone for tracking yields accumulative error, as errors of previous inferences will propagate over time to current estimation in incremental localization. Thus, refining previous results becomes necessary. Registration between estimated depth maps and pre-operative bronchi model provides absolute position of bronchoscope, eliminating accumulative error from previous relative pose inference.

With depth map registration, after generating depth map $\hat{z}_t$ from input bronchoscopic frame $x_t$, camera pose $P_t$ is estimated by minimizing the difference between $\hat{z}_t$ and rendered depth $z\left(P_t^{\prime}\right)$ in pre-operative airway model at pose $P_t^{\prime}$. The optimization process is described as:

\begin{equation}\label{eq:example}
  \begin{aligned}
P_t=\operatorname{argmax}_{P_t^{\prime}} \operatorname{NCC}\left(\hat{z}_t, z\left(P_t^{\prime}\right)\right),
\end{aligned}
\end{equation}

\noindent where $\operatorname{NCC}(\cdot)$ is normalized cross-correlation. Because of the implicit objective function, the Powell algorithm \cite{ fletcher1963rapidly} serves as the optimization strategy. Note that during optimization, constant rendering of depth maps in each iteration is the most time-consuming part.

Embedding ego-motion inference into registration, we introduce an embedded-loop localization diagram, where DD-VNB estimates the relative pose change of the bronchoscope in high frequency while the depth map registration against the pre-operative $3 \mathrm{D}$ model periodically provides absolute location for error correction in a lower speed.

Taking one registration loop as example, denote $ref$ as the reference time point. At time $ref$, iterative registration of the estimated depth map $\hat{z}_{{ref}}$ against airway model begins. As the registration running, $\hat{z}_{{ref}}$ together with estimated depth map $\hat{z}_{{ref+i}}$ are taken as a pair of input to the ego-motion loop, inferring relative camera position $P_{{ref,ref+i}}$ at time $r e f+i$ in real-time, where $i \in[1, m]$. When $P_{{ref,ref+m}}$ has been estimated, registration for $P_{{ref}}$ has completed. A more accurate pose inference $\dot{P}_{r e f+m}$ for $r e f+m$ is obtained by concatenating $P_{{ref}}$ and $P_{{ref, ref + m}}$. For the next iteration, $r e f+m$ serves as the reference time point, and $\dot{P}_{r e f+m}$ is considered as the initial value for registration. The selection of the hyperparameter $m$ is made with the aim of aligning the computational frame rate with the video capture speed.

\section{Experiment}
\subsection{Dataset}
Our experiments span phantom and patient datasets, training and testing our framework and benchmarks on each.

\textbf{Phantom Data:} Collected from a robotized bronchoscope, we have 13 video clips (640x480 res., 30fps), totaling 1000-3000 frames per clip. Training uses 8 right lung clips; testing uses 5 left lung clips. A high-precision CT scan of the lung phantom facilitates 3D reconstruction and segmentation for depth estimation and localization training.

\textbf{Patient Data:} Includes nine cases captured with an Olympus BF-6C260 bronchoscope by 10$\sim$15fps, supplemented by checkerboard videos for camera calibration. CT scans precede operations, with airway segmentation performed for model reconstruction. Training involves six cases with comprehensive airway videos (approx. 1500 frames each); testing involves three cases with trachea to lobar bronchus videos (150-200 frames each). Real and virtual bronchoscopy frames are manually aligned to ensure ground truth accuracy.

\textbf{Virtual Bronchoscopy Data:} Generated using the SOFA framework simulator, comprising 54 video clips (640x480 res., 400-1000 frames per case) with corresponding camera poses and depth maps for motion inference network training.

\subsection{Implementation}
Our training leverages the Pytorch framework on an NVIDIA RTX3090 GPU. We train the depth estimation network on phantom and patient datasets seperately. Initially, we maintain the learning rate at 0.0001 without enforcing consistency ($\tau_1=\tau_2=\eta=0$) to accommodate early-stage training variability. Parameters adjust after 10 epochs to $\tau_1=0.3$, $\tau_2=5$ and $\eta=5$ for refined training over 100 epochs using the Adam optimizer. $ \gamma$ and $\delta$ are set to 10, 5 and 1 respectively as in \cite{zhu2017unpaired} during the whole training process. The ego-motion network, trained with virtual data reflecting different camera intrinsics, undergoes 300 epochs at a learning rate of $1 \mathrm{e-5}$ with $\omega$ set to 100. Depth map registration employs Powell's algorithm with set error tolerance and convergence criteria of 0.01.

\begin{table*}[htbp]
    \renewcommand{\arraystretch}{1.5}
    \centering
    \caption{Monocular Depth Estimation Results}
    \resizebox{\textwidth}{!}{
    \begin{tabular}{c|ccccc|ccccc}
        \Xhline{1.5pt}
        \multirow{2}{*}{Methods} & \multicolumn{5}{c|}{\textbf{Phantom}} & \multicolumn{5}{c}{\textbf{Patient}} \\
        & SSIM & NCC & MAE [mm] & RMSE [mm] & Scale drift & SSIM & NCC & MAE [mm] & RMSE [mm] & Scale drift \\
        \hline
        EndoSLAM* & 0.918 & 0.822 & 4.786 $\pm$ 0.926 & 7.832 $\pm$ 0.804 & - & 0.796 & 0.738 & 8.415 $\pm$ 2.126 & 9.919 $\pm$ 1.947 & - \\

        CycleGAN & 0.914 & 0.911 & 5.347 $\pm$ 1.809 & 7.038 $\pm$ 1.974 & -0.256 $\pm$ 0.250 & 0.786 & 0.846 & 10.535 $\pm$ 4.547 & 13.260 $\pm$ 4.793 & -0.789 $\pm$ 0.781 \\

        \hline
        Ours w/o $L_{rec}$ & 0.916 & \textbf{0.919} & 4.781 $\pm$ 1.367 & 6.171 $\pm$ 1.321 & -0.278 $\pm$ 0.171 & 0.832 & \textbf{0.847} & 7.711 $\pm$ 3.278 & 9.936 $\pm$ 3.392 & -0.475 $\pm$ 0.684 \\

        Ours w/o $L_{rec_{\hat{x}}}$ & 0.923 & 0.899 & 4.491 $\pm$ 1.263 & 6.014 $\pm$ 1.380 & \textbf{-0.152 $\pm$ 0.235} & 0.851 & 0.828 & 6.455 $\pm$ 2.542 & 8.549 $\pm$ 2.663 & -0.276 $\pm$ 0.569 \\

        \hline
        Ours & \textbf{0.931} & 0.901 & \textbf{3.993 $\pm$ 0.860} & \textbf{5.727 $\pm$ 0.885} & -0.160 $\pm$ 0.162 & \textbf{0.862} & 0.83 & \textbf{5.772 $\pm$ 2.706} & \textbf{7.631 $\pm$ 3.217} & \textbf{-0.034 $\pm$ 0.445} \\
        
        \Xhline{1.5pt}
    \end{tabular}}
    \begin{tablenotes} 
            \footnotesize
            \item * denotes recovering scale before evaluation. MAE, RMSE and scale drift values are given as the mean $\pm$ standard deviation. The best performance in each block is indicated in bold. 
        \end{tablenotes}
    \label{table2}%
\end{table*}%

\begin{figure*}[tbp]
\centerline{\includegraphics[width=\textwidth]{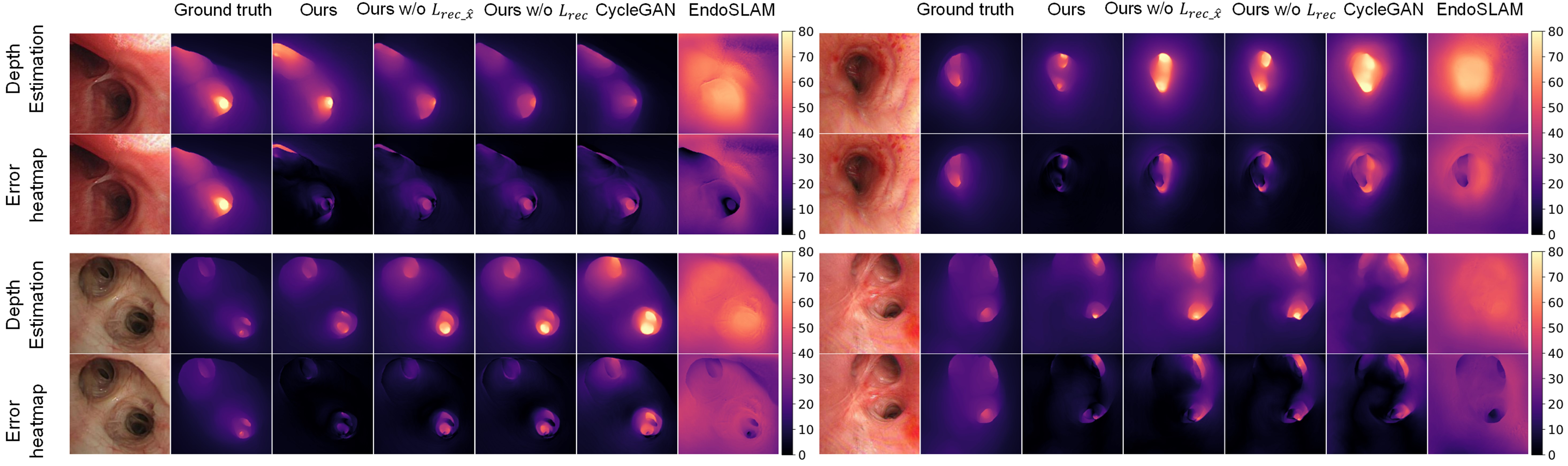}}
\caption{Quantitative depth evaluations. The original input image, depth ground truth, predicted depth maps and error heatmaps by our depth estimation, ours w/o view consistency loss for generated bronchoscopic frame, ours w/o view consistency loss, CycleGAN and EndoSLAM are shown from left to right.}
\label{fig5}
\end{figure*}

\section{Results}
\subsection{Depth Estimation Evaluation}
As no ground truth camera movement and depth map is available during training process, monocular depth estimation methods of EndoSLAM \cite{ozyoruk2021endoslam} and CycleGAN \cite{zhu2017unpaired} are employed as baselines for comparison. CycleGAN is the most commonly used depth estimation method in learning based VNB. EndoSLAM jointly trains unsupervised monocular depth estimation and camera pose networks by adopting view consistency and geometry consistency from \cite{bian2019unsupervised}.

Results are shown in Table \ref{table2}. To further validate the scale-awareness of our depth estimation, we add scale drift in the measurements, which is defined by:

\begin{equation}\label{eq:example}
{ \text{scale drift}}=1-\frac{\operatorname{mean}(\hat{z})}{\operatorname{mean}(z)}.
\end{equation}

To address scale ambiguity in EndoSLAM depth maps, we align the first frame's average depth with ground truth, excluding scale drifts from our evaluation. Our method sets a new standard in depth estimation, with a notable low RMSE of $5.727 \pm 0.885$ mm for phantom data and $7.631 \pm 3.217$ mm for patient data, outperforming CycleGAN by 18.6\% and 42.5\% in RMSE reduction for phantom and patient data, respectively. This improvement, especially against CycleGAN's baseline, underscores our method's effective use of knowledge embedding for accurate depth estimation within absolute scale. We selected several typical images from patient data for the qualitative evaluation of depth estimation methods in Fig. \ref{fig5}.

\subsection{Bronchoscopy Localization Evaluation}
All methods are trained and evaluated on phantom data and in-vivo patient data respectively. Our method and CycleGAN additionally necessitate the inclusion of virtual data captured with corresponding intrinsic parameters.

Due to the variations in video capturing frequencies, $m$ is selected to be 10 for phantom data and 3 for patient data, in order to ensure sufficient co-visibility between input frames for ego-motion estimation. We also notice enormous error when testing EndoSLAM localization performance in phantom data, due to lack of parallax between consecutive frames and the accumulative error during incremental pose inference. Instead, we test EndoSLAM by sampling one out of every ten frames from phantom data as input frames. 

Results in phantom and patient data are shown in Table \ref{table4} and Table \ref{table5} respectively. Evaluation metrics follows existing standards \cite{shen2019context,gu2022vision}. According to the tests in terms of ATE on data in phantom and patients, our method outperforms SOTA. The average ATE of our proposed framework is $4.7 \pm 3.17$ mm in phantom and $6.49 \pm 3.88$ mm in patient data, far exceeds the performance of EndoSLAM and CycleGAN with depth registration, which is the mostly reported in depth based bronchoscopic localization literature. Our method also achieved the highest SR-5 and SR-10, adding further proof for its superiority. We observe minor performance deterioration of our method from phantom data to patient data. 

\begin{figure}[tbp]
\centerline{\includegraphics[width=\columnwidth]{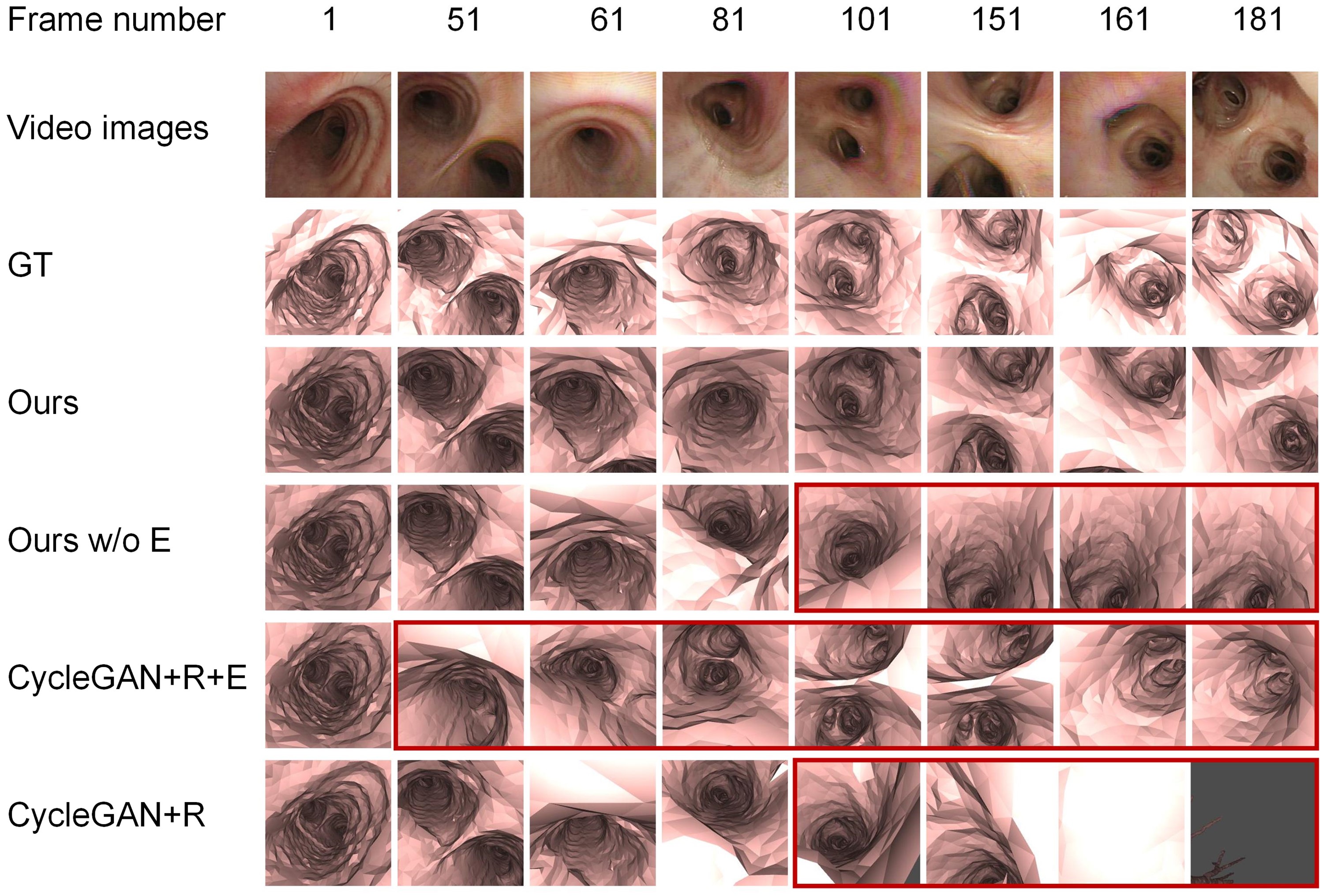}}
\caption{Example of located virtual view using different localization frame-works. E represents ego-motion estimation and R represents registration. Incremental tracking methods (including EndoSLAM, DD-VNB w/o R) are not included because most of their located views are outside the airway model. Frames where tracking was lost are box selected in red.}
\label{fig6}
\end{figure}

\begin{figure}[tbp]
\centerline{\includegraphics[width=7cm]{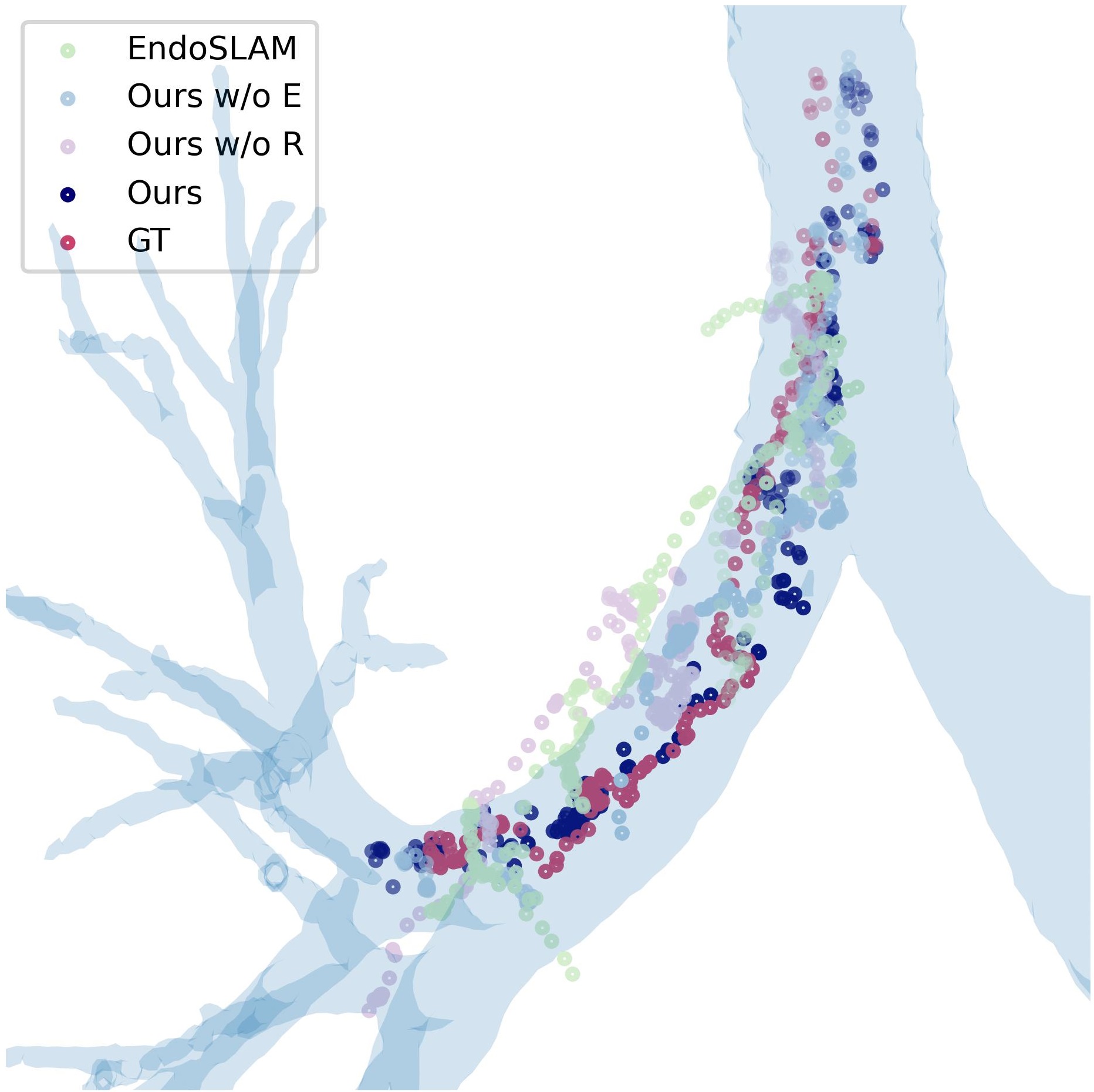}}
\caption{Tracked positions using different localization frameworks are plotted against ground truth. E represents ego-motion estimation and R represents registration. Ours shows better performance, with tracking positions closely following the ground truth trajectory.}
\label{fig7}
\end{figure}

We provide an example case (Case-1) in patient data, showing the located view in virtual bronchoscope in Fig. \ref{fig6}. Fig. \ref{fig7} shows the located position by different frameworks in Case-3 against their ground truth.

Our overall localization framework reaches average update frequency 33.9 Hz for phantom data and 12.2 Hz for patient data, due to different $m$ value. Reported learning based VNB algorithms that exceed our computation speed including \cite{sganga2019offsetnet}, \cite{sganga2019autonomous} and \cite{zhao2019generative}, set out to track camera position by a single localization network, which contributes to their high computation speed. However, their methods are difficult, if not impossible, to generalize to unseen airways or branches.

\begin{table}[tbp]
  \renewcommand{\arraystretch}{1.5}
  \setlength\tabcolsep {1pt}
  \centering
  \caption{Localization Results On Phantom}
  \resizebox{\columnwidth}{!}{
  \begin{tabular}{ccccccc}
    \toprule
        \textbf{Methods} & \textbf{ATE (mm)} & \textbf{SR-5 (\%)} & \textbf{SR-10 (\%)} & \textbf{Runtime} \\
        \midrule
        
        \textbf{EndoSLAM* (O)} & 16.68 $\pm$ 9.12 & 6.10\% & 30.60\% &         {\textbf{141Hz}} \\

        \textbf{CycleGAN+R} & 12.28 $\pm$ 11.99 & 52.80\% & 73.50\% & 3.8Hz \\
        \textbf{CycleGAN+E (O)} & 9.08 $\pm$ 5.04 & 11.40\% & 47.00\% & {43Hz} \\
        \textbf{CycleGAN+E+R (O)} & 7.45 $\pm$ 4.71 & 42.00\% & 79.60\% & {33.9Hz} \\

        \hline

        \textbf{Ours w/o R (O)} & 9.65 $\pm$ 3.96 & 9.80\% & 61.60\% & {43Hz} \\
        
        \textbf{Ours w/o E} & 7.35 $\pm$ 6.1 & 50.20\% & 70.20\% & 3.8Hz \\

        \hline
        
        \textbf{Ours (O)} & \textbf{4.7 $\pm$ 3.17} & \textbf{59.20\%} & \textbf{88.70\%} & {33.9Hz} \\
        \bottomrule
  \end{tabular}}
  \begin{tablenotes} 
        \footnotesize
        \item * denotes aligning global scale before evaluation. (O) denotes computational speed approaching video capture speed. E represents ego-motion estimation and R represents registration. ATE is given as the mean $\pm$ standard deviation. The best performance is indicated in bold. 
    \end{tablenotes}
  \label{table4}%
\end{table}%

\subsection{Ablation Studies for Consistency Losses}
In order for our depth estimation network to obtain scale perception, we have integrated view consistency losses and geometry consistency loss during CycleGAN-style network training. By those consistency constraints, we are expecting to achieve lower scale drifts in our depth estimation, thereby reducing the estimation error. Defining $L_{r e c_{-} \hat{x}}$   as the view consistency loss for generated bronchoscopic frame, $L_{rec_{-}x}$ as the view consistency loss for real bronchoscopic frame, and  $L_{gc}$ as geometry consistency loss, here we specifically investigate the following cases:

\begin{enumerate}
    \item our depth estimation with $L_{r e c_{-} \hat{x}}$ , $L_{rec_{-}x}$ and $L_{gc}$, 
    \item our depth estimation with $L_{rec_{-}x}$ and $L_{gc}$,  without $L_{r e c_{-} \hat{x}}$,
	\item our depth estimation with $L_{gc}$, without $L_{r e c_{-} \hat{x}}$ and $L_{rec_{-}x}$,
	\item our depth estimation without  $L_{rec}$ and $L_{gc}$,  deteriorates as CycleGAN.
\end{enumerate}

The results for depth estimation are given in Table \ref{table2}. As seen from quantitative ablation analysis, the consistency losses make our depth estimation more accurate in both phantom and patient data. View consistency loss which utilizes camera movement for bronchoscopic frame view synthesis, contributes the most with regard to scale perception, and in turn reduces the estimation error.

\subsection{Ablation Studies for Localization Framework}
We conduct ablation studies to assess various depth estimation and pose inference combinations within our localization framework. Table \ref{table4} and Table \ref{table5} shows the results on phantom and patient datasets respectively. Initially, we evaluate our depth estimation technique against CycleGAN, highlighting its impact on bronchoscopic localization. We further examine the role of ego-motion estimation by comparing localization accuracy with and without it, ensuring consistent searching spaces and hyperparameters across tests. Our findings underscore the significance of integrating accurate depth estimation and ego-motion inference for enhanced localization performance, particularly noting ego-motion's sensitivity to depth estimation scale drift. Results on both phantom and patient datasets confirm our framework's effectiveness, with ego-motion inference proving crucial for avoiding local minima in depth map registration.

\begin{table}[tbp]
  \renewcommand{\arraystretch}{1.5}
  \setlength\tabcolsep{1pt}
  \centering
  \caption{Localization Results On Patient Data}
  \resizebox{\columnwidth}{!}{
  \begin{tabular}{ccccccc}
    \toprule
     \textbf{Methods} & \textbf{ATE (mm)} & \textbf{SR-5 (\%)} & \textbf{SR-10 (\%)} & \textbf{Runtime} \\
    \midrule
    \textbf{EndoSLAM* (O)} & 12.48 $\pm$ 5.19 & 5.70\% & 29.00\% & {\textbf{141 Hz}} \\
   
    \textbf{CycleGAN+R} & 15.13 $\pm$ 10.07 & 21.70\% & 54.50\% & 3.8 Hz \\
    \textbf{CycleGAN+E (O)} & 15.58 $\pm$ 9.29 & 12.20\% & 58.00\% & {43 Hz} \\
    \textbf{CycleGAN+E+R (O)} & 30.8 $\pm$ 16.61 & 8.52\% & 20.95\% & {12.2 Hz} \\
    \hline
         
    \textbf{Ours w/o R (O)} & 11.13 $\pm$ 6.76 & 2.80\% & 38.00\% & {43 Hz} \\

    \textbf{Ours w/o E} & 15.97 $\pm$ 12.99 & 26.30\% & 53.80\% & 3.8 Hz \\
    \hline

    \textbf{Ours (O)} & \textbf{6.49 $\pm$ 3.88} & \textbf{47.10\%} & \textbf{85.00\%} & {12.2} Hz \\
    \bottomrule
  \end{tabular}}
  \begin{tablenotes} 
        \footnotesize
        \item * denotes aligning global scale before evaluation. (O) denotes computational speed approaching video capture speed. E represents ego-motion estimation and R represents registration. ATE is given as the mean $\pm$ standard deviation. The best performance is indicated in bold. 
    \end{tablenotes}
  \label{table5}%
\end{table}%

\subsection{Runtime}
All of the methods are tested on a workstation with a 12th Gen Intel® Core ™ i7-12700 CPU and a NVIDIA RTX3090 GPU. Input images and depth maps are all cropped and resized to 256 × 256-res. Results are presented in Table \ref{table4} and Table \ref{table5}. Runtime variance of our localization framework between phantom and patient data lies on the different $m$ value. Depth map rendering takes up the most time in registration methods. Rendering one frame takes about 1.7ms in our implementation. Rendering acceleration should further speed up our overall framework.

\section{Conclusion}
Our study proposes a bronchoscopic localization framework, featuring a knowledge-embedded depth estimation network within a dual-loop scheme for fast and accurate localization. Utilizing monocular frames for depth map inference and integrating both ego-motion estimation and airway CT registration, our work marks a stride toward real-time, learning-based bronchoscopic localization capable of adapting to unseen airways, with future enhancements aimed at further accuracy improvements through feature matching and relocalization strategies.

\AtNextBibliography{\small}
\printbibliography

\end{document}